\documentclass[letterpaper, 10 pt, conference]{ieeeconf}  
\usepackage[left=54pt, right=54.5pt, top=54.5pt, bottom=57pt]{geometry}

\IEEEoverridecommandlockouts                              

\usepackage{booktabs}
\usepackage{threeparttable}
\usepackage{cite}
\usepackage{indentfirst}
\usepackage{setspace}
\usepackage{CJK}
\usepackage{amsmath}
\usepackage{amssymb}
\usepackage{caption}
\usepackage{geometry}
\usepackage{float}
\usepackage{algorithm2e}
\usepackage{multirow}
\usepackage{pbox}

\usepackage{graphicx}
\usepackage{subcaption}
\usepackage{color}

\usepackage[symbol]{footmisc}

\title{\LARGE \bf
End-to-end Autonomous Driving Perception with Sequential Latent Representation Learning}

\author{Jianyu Chen, Zhuo Xu and Masayoshi Tomizuka
\thanks{J. Chen, Zhuo. Xu and M. Tomizuka are with Department of Mechanical Engineering, University of California, Berkeley, CA94720, USA. \quad \texttt{Email: \{jianyuchen, zhuoxu, tomizuka\}@berkeley.edu}}%
}

\begin{document}

\maketitle
\thispagestyle{empty}
\pagestyle{empty}

\begin{abstract}
Current autonomous driving systems are composed of a perception system and a decision system. Both of them are divided into multiple subsystems built up with lots of human heuristics. An end-to-end approach might clean up the system and avoid huge efforts of human engineering, as well as obtain better performance with increasing data and computation resources. Compared to the decision system, the perception system is more suitable to be designed in an end-to-end framework, since it does not require online driving exploration. In this paper, we propose a novel end-to-end approach for autonomous driving perception. A latent space is introduced to capture all relevant features useful for perception, which is learned through sequential latent representation learning. The learned end-to-end perception model is able to solve the detection, tracking, localization and mapping problems altogether with only minimum human engineering efforts and without storing any maps online. The proposed method is evaluated in a realistic urban driving simulator, with both camera image and lidar point cloud as sensor inputs. The codes and videos of this work are available at our github repo\footnote[2]{\texttt{https://github.com/cjy1992/detect-loc-map}} and project website\footnote[3]{\texttt{https://sites.google.com/berkeley.edu/e2e-percep}}.

\end{abstract}

\section{Introduction}
Motivated by the potential social impact and fueled by the recent advances in both hardware (sensor technologies such as lidar) and software (artificial intelligence techniques such as deep learning), massive efforts from both industry and academia have been invested to autonomous driving during the last decade. Start from the DARPA urban challenges~\cite{urmson2008autonomous,montemerlo2008junior}, a number of autonomous vehicle system demonstrations have been performed. Automotive industry giants such as Benz~\cite{ziegler2014making}, and IT giants such as Google~\cite{bansal2018chauffeurnet} are competing to develop the first commercial fully autonomous vehicle.

\begin{figure}
  \centering
  \includegraphics[width = .48\textwidth]{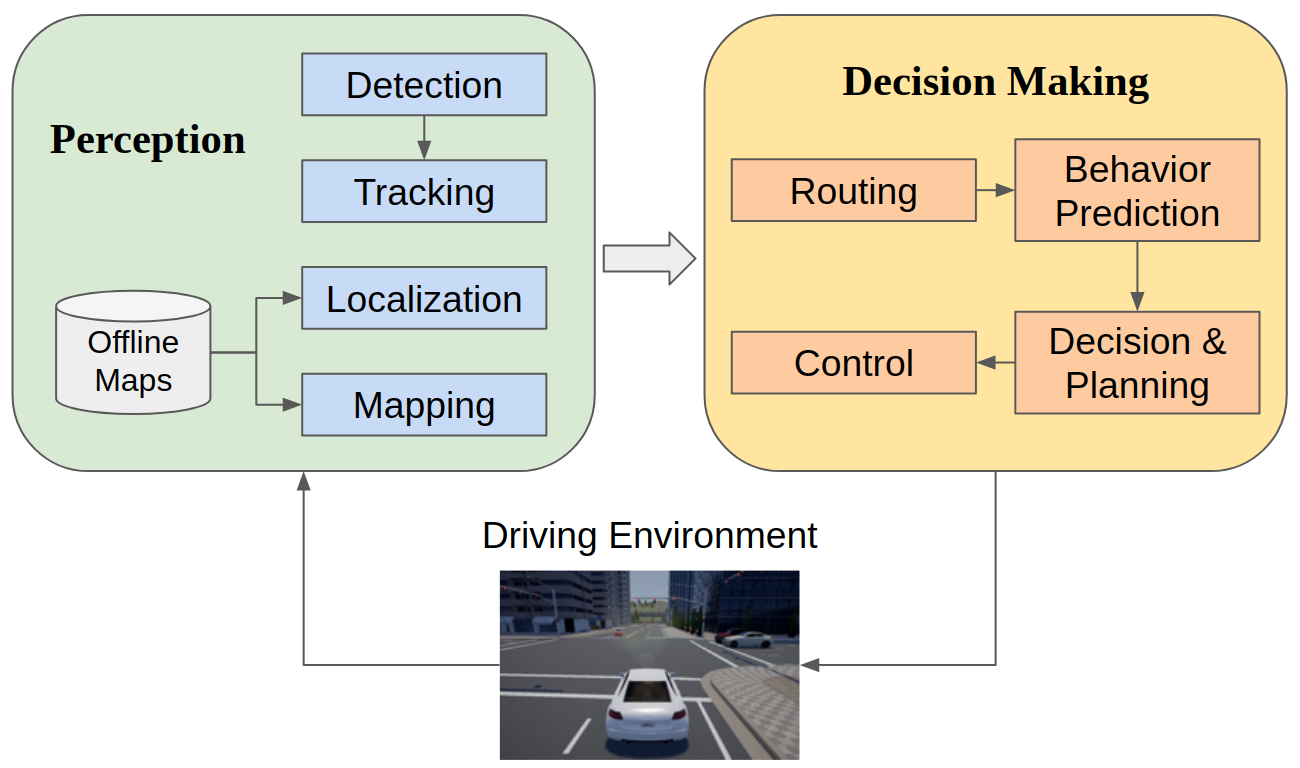}
  \caption{\label{Fig:architecture}Typical architecture of autonomous driving systems, which is composed of the perception part and decision making part with multiple subsystems.}
\end{figure}

A typical autonomous driving system is organized as a two parts architecture~\cite{badue2019self,yurtsever2019survey}: a perception system, and a decision making system, as shown in Fig.\ref{Fig:architecture}. The perception system is composed of multiple subsystems including detection~\cite{yang2018pixor}, tracking~\cite{luo2014multiple}, localization and mapping~\cite{bresson2017simultaneous}. These systems, together with offline collected maps, transform the raw sensor inputs (e.g, camera RGB images and lidar point clouds) to useful information such as surrounding vehicles' poses, ego vehicle's pose, and the local semantic map centered around the ego vehicle. On the other hand, the decision making system is divided into subsystems including routing~\cite{bauer2010combining}, behavior prediction~\cite{tang2019adaptive}, decision \& planning~\cite{chen2018continuous,chen2019autonomous}, and control~\cite{paden2016survey} that work together to generate a control command (e.g, steering angle, throttle and braking) to drive the autonomous car.

Although this highly modularized architecture works well in a few driving tasks, it starts to touch its performance limitations because (1) too much human heuristics can lead to inappropriate perception results and driving behaviors; and (2) Too many complicated subsystems are making the whole system expensive to scale and maintain. Alternatively, end-to-end architectures might avoid those limitations, as the driving models can be learned and continuously optimized from data, without much hand-engineered involvement. However, although efforts have been made to build such autonomous driving architectures~\cite{chen2020interpretable,chen2019model,chen2019deep}, an end-to-end system with decision making in the loop can only be demonstrated in simulations, real world applications are still infeasible due to safety issues. On the other hand, the perception system alone is suitable to be designed in an end-to-end form, since it does not require online driving exploration.

In this paper, we propose an end-to-end approach for the autonomous driving perception system defined in Fig.\ref{Fig:architecture}. This method enables us to solve the detection, tracking, localization and mapping problems altogether, by learning a sequential latent representation model. The learned model is able to simultaneously provide accurate estimation of surrounding vehicle poses, ego vehicle global pose, and local semantic roadmap. With this end-to-end approach, we only need minimum human engineering efforts to obtain a fully functional perception system, and no maps are needed online. Furthermore, the fusion of these subsystems helps improve the performance of surrounding vehicle pose estimation.

The remainder of this paper is organized as follows. Section~\ref{sec:related} summarizes existing works for autonomous driving perception. Section~\ref{sec:pre} analyzes the subsystems of a typical perception system to help us better understand their purposes and principles. Details of our proposed method is introduced in section~\ref{sec:proposed}. Section~\ref{sec:experiments} shows the experiments and results, while section~\ref{sec:conclusion} concludes the paper.

\section{Related Work}\label{sec:related}
It is crucial to perceive the environment and extract useful information. This mainly includes vehicle detection, tracking, localization and mapping.
\subsection{Vehicle Detection}
Vehicle detection refers to estimating the position, heading and size of surrounding vehicles. There are three main subclasses: (1) Image-based vehicle detection generates bounding boxes on front-view camera image~\cite{redmon2016you,ren2015faster}; (2) Semantic segmentation assigns each pixel of the image with a class label. Pixels belonging to same objects have the same label~\cite{long2015fully,he2017mask}; and (3) 3D object detection obtains the 3D poses (or bird-eye poses) of surrounding vehicles, which is usually achieved with the help of lidar point cloud~\cite{chen2017multi,yan2018second}.

\subsection{Vehicle Tracking}
Direct prediction from the vehicle detection system is often insufficient, more accurate vehicle state needs to be estimated given historical detection. Typical vehicle tracking methods have two phases: (1) data association to connect objects between frames~\cite{hwang2016fast,nguyen2011stereo}; and (2) filtering methods such as Kalman Filters and Particle Filters to smooth the vehicle dynamics~\cite{ess2010object,petrovskaya2009model}.

\subsection{Localization and Mapping}
Localization is the task of estimating ego vehicle pose relative to a reference frame in a map, which can be either a raw point cloud map or an annotated semantic map, depending on the algorithm we choose. There are two main approaches: (1) Simultaneous localization and mapping (SLAM) makes map online and localize the ego vehicle in the map at the same time~\cite{bresson2017simultaneous}; and (2) A priori map-based localization that estimate the ego vehicle pose by finding the best match to a detailed a priori map~\cite{levinson2007map}.

\section{Preliminary}\label{sec:pre}
In this section, we will analyze the typical subsystems in an autonomous driving perception system, including detection, tracking, localization and mapping. They are reformulated into graphical models. This helps us better understand their purposes and relationships to fuse them into a single end-to-end framework.

\subsection{Typical Detection and Tracking Systems}
The goal of the detection and tracking subsystems is to accurately estimate the states of surrounding vehicles. This includes their positions and heading angles relative to the ego vehicle, as well as their length and width. Current detection and tracking systems are divided into two subsystems. First, the detection subsystem predicts the vehicles' poses based on single frame sensor inputs. Then, the tracking subsystem smooths the results from the detection system based on its historical outputs.

\begin{figure}
  \centering
  \includegraphics[width = .42\textwidth]{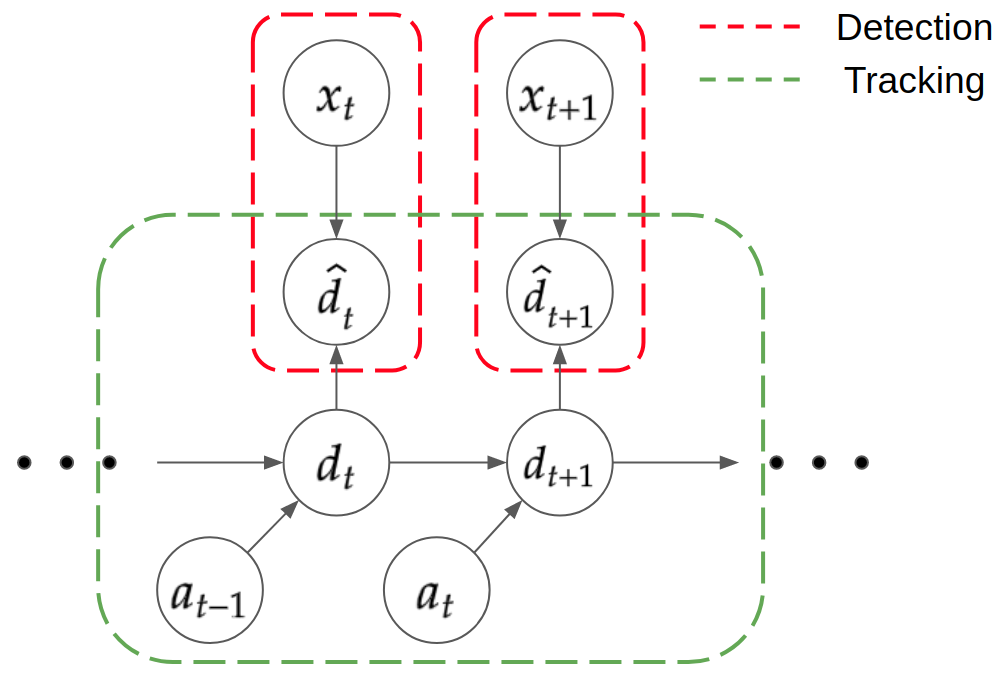}
  \caption{\label{Fig:detection}Graphical model of typical detection and tracking systems. The red block represents the detection system. The green block represents the tracking system.}
\end{figure}

These processes can be interpreted as a graphical model, as shown in Fig~\ref{Fig:detection}. The red block represents the detection subsystem, which contains a detection model $p\left(\hat{d}_t|x_t\right)$ that maps the sensor inputs $x_t$ to an estimation of the surrounding vehicles' poses $\hat{d}_t$. This part is usually performed by fitting a deep neural network model using supervised learning techniques. The green block represents the tracking subsystem, which estimates the true vehicles' poses $d_t$ given historical outputs of the detection system $\hat{d}_{1:t}$ and sometimes the historical ego vehicle actions $a_{1:t}$. This estimation problem can be formulated as a filter problem $p\left(d_t|\hat{d}_{1:t},a_{1:t}\right)$ which is usually solved by Kalman filters with hard-coded transition model $p\left(d_{t+1}|d_t,a_t\right)$ and observation model $p\left(\hat{d}_t|d_t\right)$.

\subsection{Typical Localization and Mapping Systems}
The localization and mapping system needs to accurately estimate the pose of the ego vehicle in the coordinate of a global map. Then a local semantic map indicating road geometry, topology and traffics will be obtained, which are used for downstream planning and control tasks. The estimated global ego vehicle pose is also used for routing.

\begin{figure}
  \centering
  \includegraphics[width = .43\textwidth]{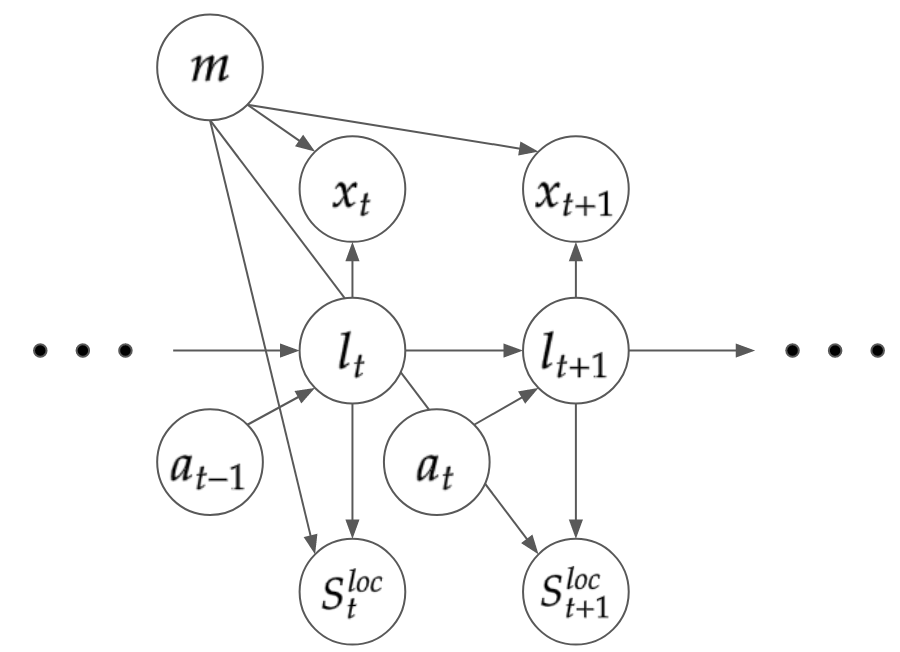}
  \caption{\label{Fig:slam}Graphical model of typical localization and mapping system.}
\end{figure}

These processes can be interpreted as a graphical model, as shown in Fig~\ref{Fig:slam}. The global ego vehicle pose $l_t$ is estimated based on historical sensor inputs and actions $p\left(l_t|x_{1:t}, a_{1:t}\right)$. The global map $m$ can be edited offline or constructed online, depending on the methodology we use. After estimating the ego vehicle pose, a local semantic map $S^{loc}_t$ is obtained by locating the ego vehicle on the global map, which requires semantic annotations. Note that the global map $m$ needs to be stored online.

\section{Proposed Method}\label{sec:proposed}
In this paper, we propose a novel end-to-end autonomous driving perception system for simultaneous detection, tracking, localization and mapping. All the functionalities are fused in a single framework. A sequential latent representation learning process is performed to learn this end-to-end perception model.

\subsection{End-to-end Autonomous Driving Perception}
In general, the purpose of detection and tracking is to estimate surrounding vehicles' poses $d_t$, while the purpose of localization and mapping is to obtain the local semantic map $S^{loc}_t$ and the global ego vehicle pose $l_t$. All these estimations are conditioned on historical sensor inputs $x_{1:t}$ and actions $a_{1:t}$. Therefore, the tasks of perception can be simplified by estimating the following conditional probability:
\begin{equation}\label{Eq:task}
    p\left(d_t,S_t^{loc},l_t|x_{1:t},a_{1:t}\right)
\end{equation}

\begin{figure}
  \centering
  \includegraphics[width = .45\textwidth]{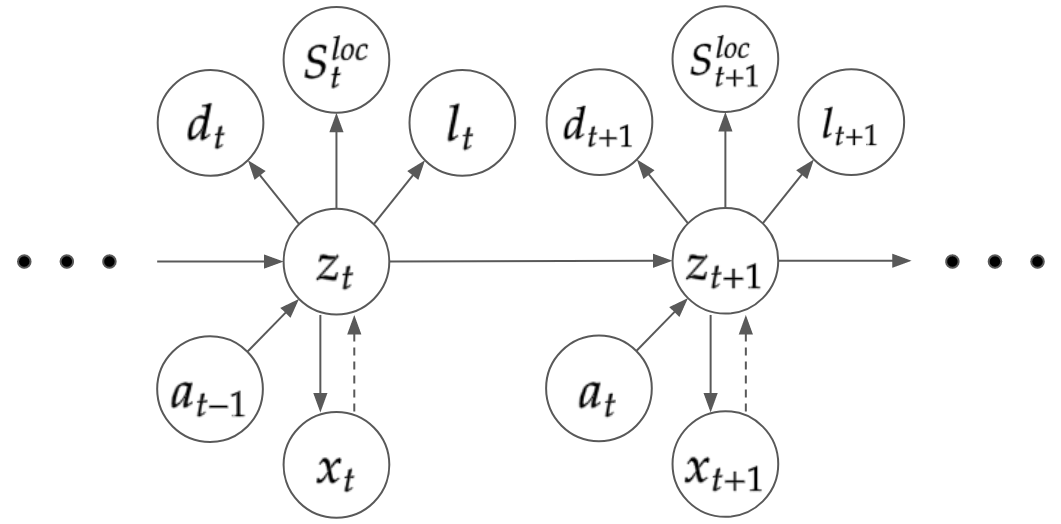}
  \caption{\label{Fig:e2e}Graphical model of end-to-end autonomous driving perception.}
\end{figure}

\begin{figure*}
  \centering
  \includegraphics[width = .95\textwidth]{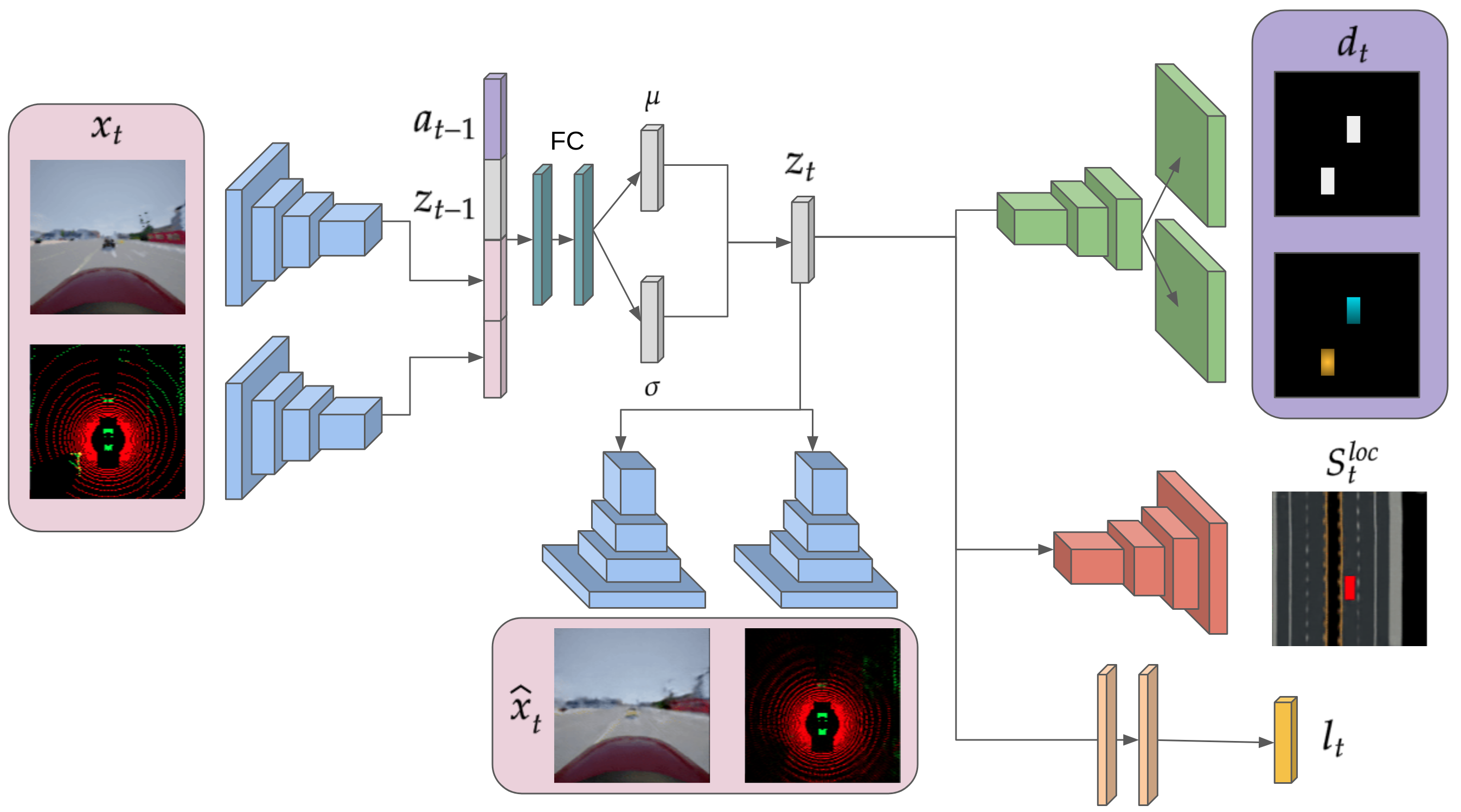}
  \caption{\label{Fig:arch}Architecture of our proposed end-to-end perception model at a single frame.}
\end{figure*}

As stated in section~\ref{sec:pre}, typical methods estimate~\eqref{Eq:task} by dividing it into multiple separate tasks, and tackle them one-by-one with lots of human engineering efforts. Different from typical methods, we propose to solve the problem of estimating~\eqref{Eq:task} jointly. This is possible by assuming that there is a latent space summarizing all useful historical information. Then with this latent space, we can extract the information we need, such as surrounding vehicles' poses, road geometry, and ego vehicle pose. Inspired by works that learn latent representations with time sequence reasoning~\cite{krishnan2015deep,lee2019stochastic,chen2020interpretable}, we propose to formulate the end-to-end perception system as a single graphical model, as shown in Fig.~\ref{Fig:e2e}. A more detailed architecture of our model in a single frame is shown in Fig.~\ref{Fig:arch}

$z_t$ is the latent variable we introduce, it represents a summary of historical information and is evolved with the latent dynamics $p\left(z_{t+1}|z_t,a_t\right)$. All relevant information about the environment are decoded from this latent, including the sensor inputs $p\left(x_t|z_t\right)$, surrounding vehicles' poses $p\left(d_t|z_t\right)$, local semantic roadmap $p\left(S_t^{loc}|z_t\right)$ and global ego vehicle pose $p\left(l_t|z_t\right)$. If we are able to estimate the distribution of the latent state given historical sensor inputs and actions $q\left(z_t|x_{1:t},a_{1:t}\right)$, then \eqref{Eq:task} can be obtained by integrating out the latent state:
\begin{equation}
\begin{aligned}
&p\left(d_t,S_t^{loc},l_t|x_{1:t},a_{1:t} \right)\\
&\qquad\qquad=\int p\left(d_t,S_t^{loc},l_t|z_t \right) q\left(z_t|x_{1:t},a_{1:t}\right) d z_t
\end{aligned}
\end{equation}

Generally we do not need to calculate the exact integration, but rather its expectation, which can be approximated by sampling:
\begin{equation}
\hat{z}_t \sim q\left(z_t|x_{1:t},a_{1:t}\right) \quad \hat{d}_t,\hat{S}_t^{loc},\hat{l}_t \sim p\left(d_t,S_t^{loc},l_t|\hat{z}_t \right)
\end{equation}

The next subsection will introduce how we can learn a model to estimate $q\left(z_t|x_{1:t},a_{1:t}\right)$ and $p\left(d_t,S_t^{loc},l_t|z_t \right)$ with sequential latent representation learning.

\subsection{Sequential Latent Representation Learning}
To learn appropriate models shown in Fig.\ref{Fig:e2e}, we need to fit them with collected dataset. For convenience, we first denote a trajectory to be composed of sensor inputs, detection outputs, local semantic roadmaps, ego vehicle poses and actions:
\begin{equation}
    \vec{x}=x_{1:t},\,\vec{d}=d_{1:t},\,\vec{S^{loc}}=S^{loc}_{1:t},\,\vec{l}=l_{1:t},\,\vec{a}=a_{1:t}
\end{equation}

The dataset is then composed of this kind of trajectories collected while driving ${\cal D} = \left \{ \left(\vec{x}^i, \vec{d}^i, \left(\vec{S^{loc}}\right)^i, \vec{l}^i, \vec{a}^i \right ) \right \}_{i=1}^N$. The model can be fitted by maximizing the log likelihood of the data:
\begin{equation}
\begin{aligned}
&\textrm{log}\, \prod_{i=1}^N p\left (\vec{x}^i, \vec{d}^i, \left(\vec{S^{loc}}\right)^i, \vec{l}^i \left|\right. \vec{a}^i\right ) \\
&\qquad \qquad \qquad =\sum_{i=1}^N \textrm{log}\,p\left (\vec{x}^i, \vec{d}^i, \left(\vec{S^{loc}}\right)^i, \vec{l}^i \left|\right. \vec{a}^i\right )
\end{aligned}
\end{equation}
which can be maximized using stochastic gradient descent (SGD), which optimizes parametric functions with gradient descent. The gradient is estimated by sampling a batch of data points. To apply SGD to our problem, $p\left (\vec{x}, \vec{d}, \vec{S^{loc}}, \vec{l} \left|\right. \vec{a}\right )$ needs to be composed of parametric functions, thus auto-differentiation tools such as TensorFlow can be used to evaluate their gradients. Variational inference~\cite{kingma2013auto} can be applied to compute this log likelihood. First, introduce the latent variables $\vec{z}=z_{1:t}$:
\begin{equation}\label{Eq:latent}
\textrm{log}\,p\left (\vec{x}, \vec{d}, \vec{S^{loc}}, \vec{l} \left|\right.\vec{a}\right ) =\textrm{log}\,\int p\left(\vec{x}, \vec{d}, \vec{S^{loc}}, \vec{l}, \vec{z} \left|\right.\vec{a}\right) d \vec{z}
\end{equation}

Then introduce a variational distribution $q\left(\vec{z}|\vec{x},\vec{a}\right)$ into \eqref{Eq:latent}:
\begin{equation}\label{Eq:vardist}
\begin{aligned}
&\textrm{log}\,p\left (\vec{x}, \vec{d}, \vec{S^{loc}}, \vec{l} \left|\right.\vec{a}\right ) \\
&\qquad \qquad =\textrm{log}\,\int p\left(\vec{x}, \vec{d}, \vec{S^{loc}}, \vec{l}, \vec{z} \left|\right.\vec{a}\right) \frac{q\left(\vec{z}|\vec{x},\vec{a}\right)}{q\left(\vec{z}|\vec{x},\vec{a}\right)} d \vec{z}
\end{aligned}
\end{equation}

Now eliminate the integration in \eqref{Eq:vardist} by introducing expectation, and then apply Jensen's inequality:
\begin{equation}\label{Eq:ELBO}
\begin{aligned}
\textrm{log}\,p\left(\vec{x}, \vec{d}, \vec{S^{loc}}, \vec{l}|\vec{a} \right)
&=\textrm{log}\underset{q\left(\vec{z}|\vec{x},\vec{a}\right)}{\mathbb{E}}\left[\frac{p\left(\vec{x}, \vec{d}, \vec{S^{loc}}, \vec{l}, \vec{z}|\vec{a} \right)}{q\left(\vec{z}|\vec{x},\vec{a}\right)}\right]\\
&\geq \underset{q\left(\vec{z}|\vec{x},\vec{a}\right)}{\mathbb{E}}\left[\textrm{log}\,p\left(\vec{x}, \vec{d}, \vec{S^{loc}}, \vec{l}, \vec{z}|\vec{a} \right)\right.\\
&\qquad \qquad \qquad \qquad \left.-\textrm{log}\,q\left(\vec{z}|\vec{x},\vec{a}\right)\right]\\
&=\text{ELBO}
\end{aligned}
\end{equation}
where ELBO stands for evidence lower bound. The original log likelihood can be maximized by maximizing this ELBO. Now derive $p\left(\vec{x},\vec{d},\vec{S^{loc}},\vec{l},\vec{z}|\vec{a} \right)$ by probability factorization according to the PGM in Fig.\ref{Fig:e2e}:
\begin{equation}\label{Eq:factor}
\begin{aligned}
&p\left(\vec{x},\vec{d},\vec{S^{loc}},\vec{l},\vec{z}|\vec{a} \right)\\
&\qquad \qquad=p\left(\vec{x},\vec{d},\vec{S^{loc}},\vec{l}|\vec{z},\vec{a} \right)p\left(\vec{z}|\vec{a}\right)\\
&\qquad \qquad=p\left(\vec{x}|\vec{z}\right)p\left(\vec{d}|\vec{z}\right)p\left(\vec{S^{loc}}|\vec{z}\right)p\left(\vec{l}|\vec{z}\right)p\left(\vec{z}|\vec{a}\right)
\end{aligned}
\end{equation}

And substitute \eqref{Eq:factor} into \eqref{Eq:ELBO}, we have:
\begin{equation}\label{Eq:modellearning}
\begin{aligned}
&\textrm{ELBO}=\underset{q\left(\vec{z}|\vec{x},\vec{a}\right)}{\mathbb{E}}[\textrm{log}\,p\left(\vec{x}|\vec{z}\right)+\textrm{log}\,p\left(\vec{d}|\vec{z}\right)+\textrm{log}\,p\left(\vec{S^{loc}}|\vec{z}\right)\\
&\qquad \qquad \qquad +\textrm{log}\,p\left(\vec{l}|\vec{z}\right)+\textrm{log}\,p\left(\vec{z}|\vec{a}\right) -\textrm{log}\,q\left(\vec{z}|\vec{x},\vec{a}\right)]
\end{aligned}
\end{equation}

Now derive the components in \eqref{Eq:modellearning} by unfolding them with time. Considering the conditional dependence of PGM in Fig.\ref{Fig:e2e}. The decoding models can be unfolded as:
\begin{equation}
\begin{aligned}
&\textrm{log}\,p\left(\vec{x}|\vec{z}\right)=\textrm{log}\prod_{\tau=1}^{t}p\left(x_\tau|z_\tau\right)=\sum_{\tau=1}^{t}\textrm{log}\,p\left(x_\tau|z_\tau\right)\\
&\textrm{log}\,p\left(\vec{d}|\vec{z}\right)=\textrm{log}\prod_{\tau=1}^{t}p\left(d_\tau|z_\tau\right)=\sum_{\tau=1}^{t}\textrm{log}\,p\left(d_\tau|z_\tau\right)\\
&\textrm{log}\,p\left(\vec{S^{loc}}|\vec{z}\right)=\textrm{log}\prod_{\tau=1}^{t}p\left(S^{loc}_\tau|z_\tau\right)=\sum_{\tau=1}^{t}\textrm{log}\,p\left(S^{loc}_\tau|z_\tau\right)\\
&\textrm{log}\,p\left(\vec{l}|\vec{z}\right)=\textrm{log}\prod_{\tau=1}^{t}p\left(l_\tau|z_\tau\right)=\sum_{\tau=1}^{t}\textrm{log}\,p\left(l_\tau|z_\tau\right)\\
\end{aligned}
\end{equation}

The prior model can be unfolded using the latent state transition function:
\begin{equation}
\begin{aligned}
\textrm{log}\,p\left(\vec{z}|\vec{a}\right)
&=\textrm{log}\left[p\left(z_1\right)\prod_{\tau=1}^{t-1} p\left(z_{\tau+1}|z_\tau,a_\tau\right)\right]\\
&=\textrm{log}\,p\left(z_1\right)+\sum_{\tau=1}^{t-1} \textrm{log}\,p\left(z_{\tau+1}|z_\tau,a_\tau\right)
\end{aligned}
\end{equation}

The latent state inference model can be unfolded as:
\begin{equation}
\begin{aligned}
\textrm{log}\,q\left(\vec{z}|\vec{x},\vec{a}\right)
&=\textrm{log}\left[q\left(z_1|\vec{x},\vec{a}\right)\prod_{\tau=1}^{t-1} q\left(z_{\tau+1}|z_\tau,\vec{x},\vec{a}\right)\right]\\
&\approx \textrm{log}\left[q\left(z_1|x_1\right)\prod_{\tau=1}^{t-1} q\left(z_{\tau+1}|z_\tau,x_{\tau+1},a_\tau\right)\right]\\
&=\textrm{log}\,q\left(z_1|x_1\right)+\sum_{\tau=1}^{t-1}\textrm{log}\, q\left(z_{\tau+1}|z_\tau,x_{\tau+1},a_\tau\right)
\end{aligned}
\end{equation}

Note here we approximate $q\left(\vec{z}|\vec{x},\vec{a}\right)$ and $q\left(z_{\tau+1}|z_\tau,\vec{x},\vec{a}\right)$ with $q\left(z_1|x_1\right)$ and $q\left(z_{\tau+1}|z_\tau,x_{\tau+1},a_\tau\right)$ for simplicity. To obtain the exact values, bi-directional recurrent neural networks should be used to obtain the posterior probabilities conditioned on the whole trajectory sequence $\left(\vec{x},\,\vec{a}\right)$~\cite{krishnan2015deep}.

Now we can unfold \eqref{Eq:modellearning} with time:
\begin{equation}\label{Eq:modelELBO}
\begin{aligned}
&\textrm{ELBO} \approx \underset{q\left(\vec{z}|\vec{x},\vec{a}\right)}{\mathbb{E}}\left[\sum_{\tau=1}^t\textrm{log}\,p\left(x_\tau|z_\tau\right)+\sum_{\tau=1}^t\textrm{log}\,p\left(d_\tau|z_\tau\right)\right.\\
&\qquad+ \sum_{\tau=1}^t\textrm{log}\,p\left(S^{loc}_\tau|z_\tau\right) +\sum_{\tau=1}^t\textrm{log}\,p\left(l_\tau|z_\tau\right)\\
&\qquad-\text{D}_{\text{KL}}\left(q\left(z_1|x_1\right)||p\left(z_1\right)\right)\\
&\qquad \left.-\sum_{\tau=1}^t\text{D}_{\text{KL}}\left(q\left(z_{\tau+1}|z_\tau,x_{\tau+1},a_\tau\right)||p\left(z_{\tau+1}|z_\tau,a_\tau\right)\right)\right]
\end{aligned}
\end{equation}
and the ELBO is now decomposed to several simple networks, which will be illustrated in section~\ref{sec:architecture}.

\subsection{Input Representations}
\subsubsection{\bf{Sensor Input $x$}}
\begin{figure}
  \centering
  \includegraphics[width = .45\textwidth]{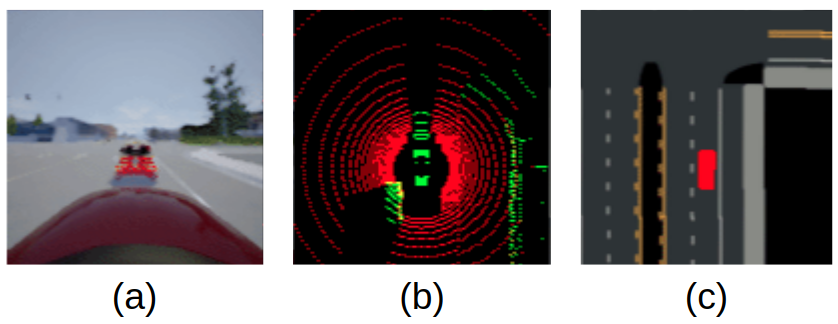}
  \caption{\label{Fig:input}Input representations. (a): Front camera RGB image; (b) lidar birdeye image; (c) local semantic roadmap. All images are with shape $128\times 128$.}
\end{figure}

We use two sensors to provide the observations, camera and lidar. For camera, the sensor input is a front-view RGB image, which can be represented by a tensor of $[0, 255]^{128\times 128\times 3}$, as shown in Fig.\ref{Fig:input}(a). For lidar, we project the point clouds to the ground plane and render them into a 2D lidar image. The lidar image is represented by a tensor of $[0, 255]^{128\times 128\times 3}$, with each pixel rendered in red or green depending on whether there are lidar points at or above ground level existing in the corresponding pixel cell, as shown in Fig.\ref{Fig:input}(b).

We use camera and lidar together because they are both important sensor sources and provide complementary information. Lidar point clouds provides accurate spatial information of other road participants and obstacles in 360 degrees of view. While the front-view camera is good at providing information of the road conditions.

\subsubsection{\bf{Detection Mask $d$}}
The detection mask is composed of two branches: a classification branch outputing a 1-channel classification feature map and a regression branch outputing a 6-channel regression feature map. Both feature maps is in bird-eye view and their representation follows that in~\cite{yang2018pixor}. Briefly speaking, the classification feature map indicates the probability of each pixel belonging to a surrounding vehicle. While the regression feature map is composed of geometry information of the corresponding surrounding vehicle, e.g, x-y positions, heading, width and length of each detected vehicle (See Section 3.1.1 in \cite{yang2018pixor} for details). With these two feature maps, we can use Non-Maximum Suppression (NMS) to get the final detection results. Thus $d$ is composed of two tensors of $\mathbb{R}^{128\times 128\times 1}$ and $\mathbb{R}^{128\times 128\times 6}$ respectively.

\subsubsection{\bf{Local Semantic Roadmap $S^{loc}$}}
The local semantic roadmap is a map centered around the ego vehicle which includes semantic information about the road geometry, road topology and traffic rules. For example, the lane markings, drivable areas, and stop signs. It is an RGB image of $[0, 255]^{128\times 128\times 3}$, as shown in Fig.\ref{Fig:input}(c).

\subsubsection{\bf{Global Ego Vehicle State $l$}}
The global ego vehicle pose includes the information of ego vehicle's x-y position (in meter) and heading angle (in rad) in the global map's coordinate. It is a vector of $\mathbb{R}^{3}$.

\subsection{Network Architectures}\label{sec:architecture}
In this section, we will illustrate the detailed architectures of the networks in \eqref{Eq:modelELBO}.
\subsubsection{\bf{Sequential Latent Model}}
\begin{figure}
  \centering
  \includegraphics[width = .48\textwidth]{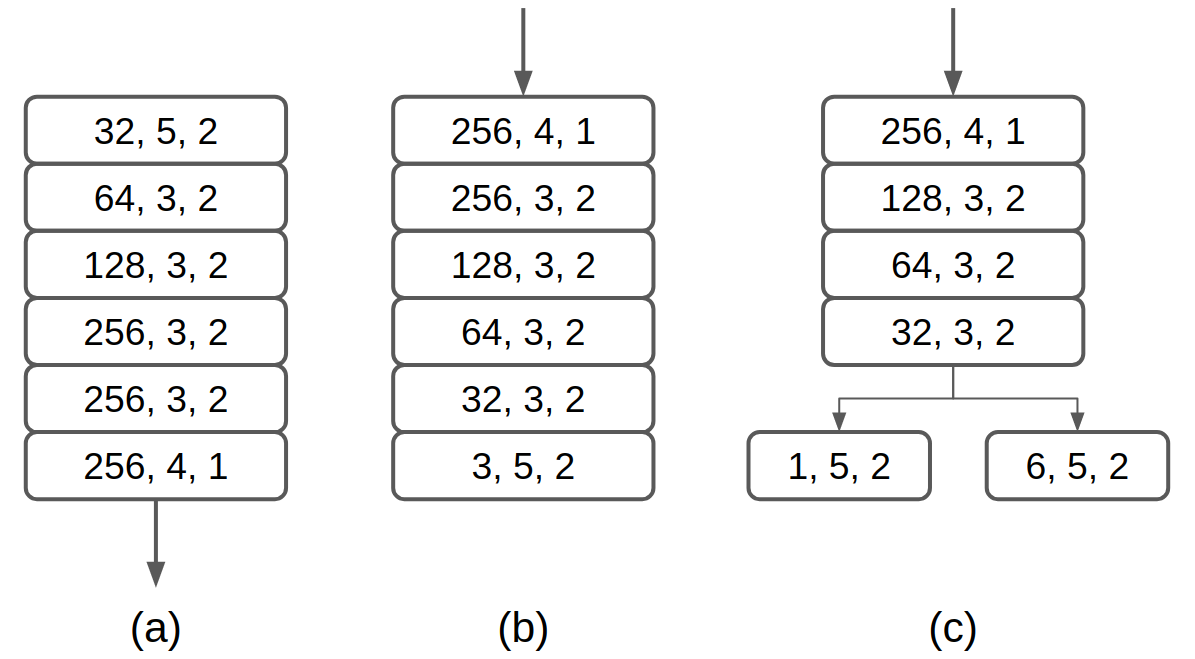}
  \caption{\label{Fig:networks}Network architectures. (a): Image encoder for camera and lidar images; (b) Image decoder for camera, lidar and roadmap images; (c) Decoder for detection masks.}
\end{figure}

The sequential latent model includes the latent dynamics network $p\left(z_{\tau+1}|z_\tau,a_\tau\right)$, the filtering model network $q\left(z_{\tau+1}|z_\tau,x_{\tau+1},a_\tau\right)$, $q\left(z_1|x_1\right)$, and the sensor inputs reconstruction network $p\left(x_\tau|z_\tau\right)$. Here we follow the two-layer hierarchical latent space structure as in \cite{lee2019stochastic}, such that $z_\tau=[z_\tau^1, z_\tau^2] \in \mathbb{R}^{288}$ where $z_\tau^1 \in \mathbb{R}^{32}$ and $z_\tau^2 \in \mathbb{R}^{256}$. $p\left(z_{\tau+1}|z_\tau,a_\tau\right)$ consists of two fully connected layers with hidden units number 256, followed by a Gaussian output layer. $q\left(z_{\tau+1}|z_\tau,x_{\tau+1},a_\tau\right)$ and $q\left(z_1|x_1\right)$ both consist of 5 convolutional layers ((32, 5, 2), (64, 3, 2), (128, 3, 2), (256, 3, 2), (256, 3, 2) and (256, 4, 1), with each tuple means (filters, kernel size, strides), as shown in Fig.\ref{Fig:networks}(a)) to first encode the sensor inputs $x_t$ into features of size 256. Then two fully connected layers with hidden units number 256 are followed, with a Gaussian output layer. $p\left(x_\tau|z_\tau\right)$ both consist of 5 deconvolutional layers ((256, 4, 1), (256, 3, 2), (128, 3, 2), (64, 3, 2), (32, 3, 2), and (3, 5, 2), with each tuple means (filters, kernel size, strides), as shown in Fig.\ref{Fig:networks}(b)) with a fixed standard deviation of 0.1. 

\subsubsection{\bf{Detection and Tracking Network}}
The detection and tracking network is the network generating the detection mask $p\left(d_\tau|z_\tau\right)$. We deployed network architecture that is similar to the network architecture in \cite{yang2018pixor}. The mask, including a one-channel tensor and a six-channel, is decoded from the latent by 4 deconvolutional layers ((256, 4, 1), (128, 3, 2), (64, 3, 2), (32, 3, 2), with each tuple means (filters, kernel size, strides)), and then a deconvolutional layer of (1, 5, 2) and (6, 5, 2) respectively, as shown in Fig.\ref{Fig:networks}(c).

\subsubsection{\bf{Localization and Mapping Network}}
The localization and mapping network includes the local semantic roadmap decoder $p\left(S^{loc}_\tau|z_\tau\right)$ and the global ego vehicle state decoder $p\left(l_\tau|z_\tau\right)$. $p\left(S^{loc}_\tau|z_\tau\right)$ has the same architecture with $p\left(x_\tau|z_\tau\right)$, as shown in Fig.\ref{Fig:networks}(b). $p\left(l_\tau|z_\tau\right)$ is a two-layer fully connected neural network with hidden units number 256.

\section{Experiments}\label{sec:experiments}
\subsection{Simulation Setup and Data Collection}\label{sec:simulation}
\begin{figure}
    \centering
    \begin{subfigure}[b]{0.22\textwidth}
        \centering
        \includegraphics[height = \linewidth]{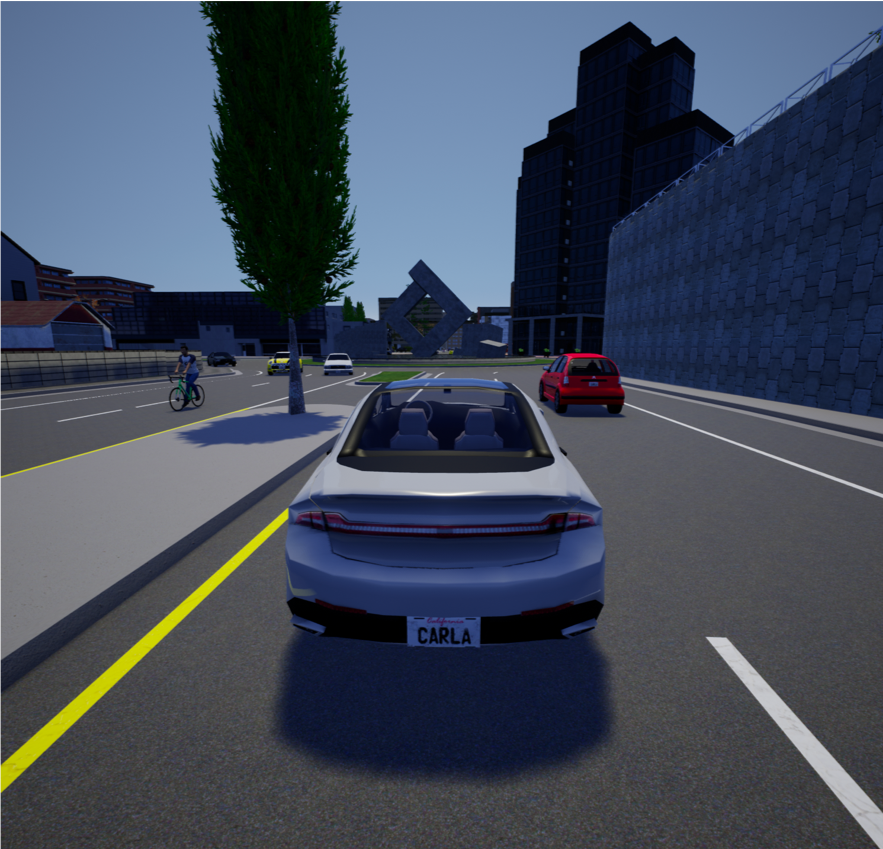}
        \caption{Sample view of CARLA simulator}
    \end{subfigure}
    ~~
    \begin{subfigure}[b]{0.22\textwidth}
        \centering
        \includegraphics[height = \linewidth]{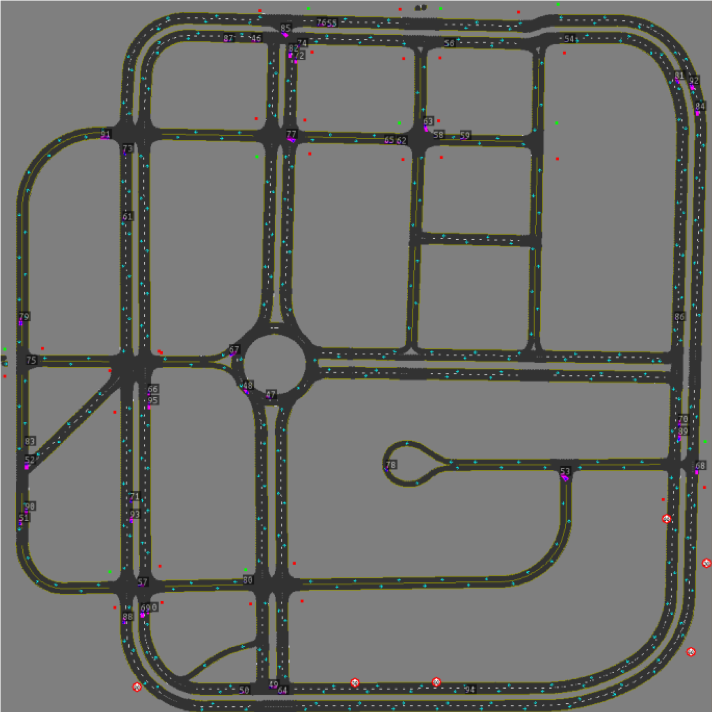}
        \caption{Map layout of the simulated city}
    \end{subfigure}
    \caption{\label{Fig:carla}Simulation environment}
\end{figure}

We train and evaluate our proposed method on CARLA~\cite{dosovitskiy2017carla}. CARLA simulator is a high-definition open-source simulation platform designed for autonomous driving research. It simulates not only the driving environment and vehicle dynamics, but also the raw sensor data inputs such as camera RGB images and lidar point clouds. Fig.\ref{Fig:carla} (a) shows a sample view of the driving simulation environment we use.

To collect the data for training, we navigate the ego vehicle in virtual town of CARLA. Fig.\ref{Fig:carla} (b) shows the map layout of the town. It includes various urban scenarios such as intersections and roundabouts. The range of the map is $400m \times 400m$, with about $6km$ total length of roads. 100 vehicles are running autonomously in the virtual town to simulate a crowded urban environment. Both the ego vehicle and the surrounding vehicles will randomly choose a direction at intersections, then follow the route, while slowing down for front vehicles and stopping when the front traffic light becomes red. We run the ego vehicle for 50k environment steps and store the observations of each step into the training dataset.

\subsection{Training Details}
The model is trained with a batch size of 32 and learning rate 0.0001. The length of sequential model used for training is $t = 10$. The total iteration of training is 100k. We train three variants of our methods:
\subsubsection{\bf{Inputs and roadmap}}
Both the raw sensor inputs and the local semantic roadmap are reconstructed when training. So the model is enforced to capture the features of both sensor inputs and the local semantic roadmap.
\subsubsection{\bf{No inputs reconstruction}}
No raw sensor inputs are reconstructed. This is reasonable since we do not necessary need to output the reconstructed raw sensor inputs.
\subsubsection{\bf{No roadmap reconstruction}}
No local semantic roadmaps are reconstructed. Then the model is not enforced to capture the features of the road geometry and traffic rules.

\subsection{Evaluation Results}\label{sec:eval}
To evaluate our method, instead of evaluating on a collected test dataset, we directly put the ego vehicle in a random start point in the virtual town with surrounding vehicles, as described in \ref{sec:simulation}. During the navigation, the ego vehicle is collecting sensor inputs, encoding it to latent states, and decoding to perception outputs. We let the vehicle runs for 15k environment steps and evaluate the performance during this period of navigation. 

\begin{figure*}
  \centering
  \includegraphics[width = 1.0\textwidth]{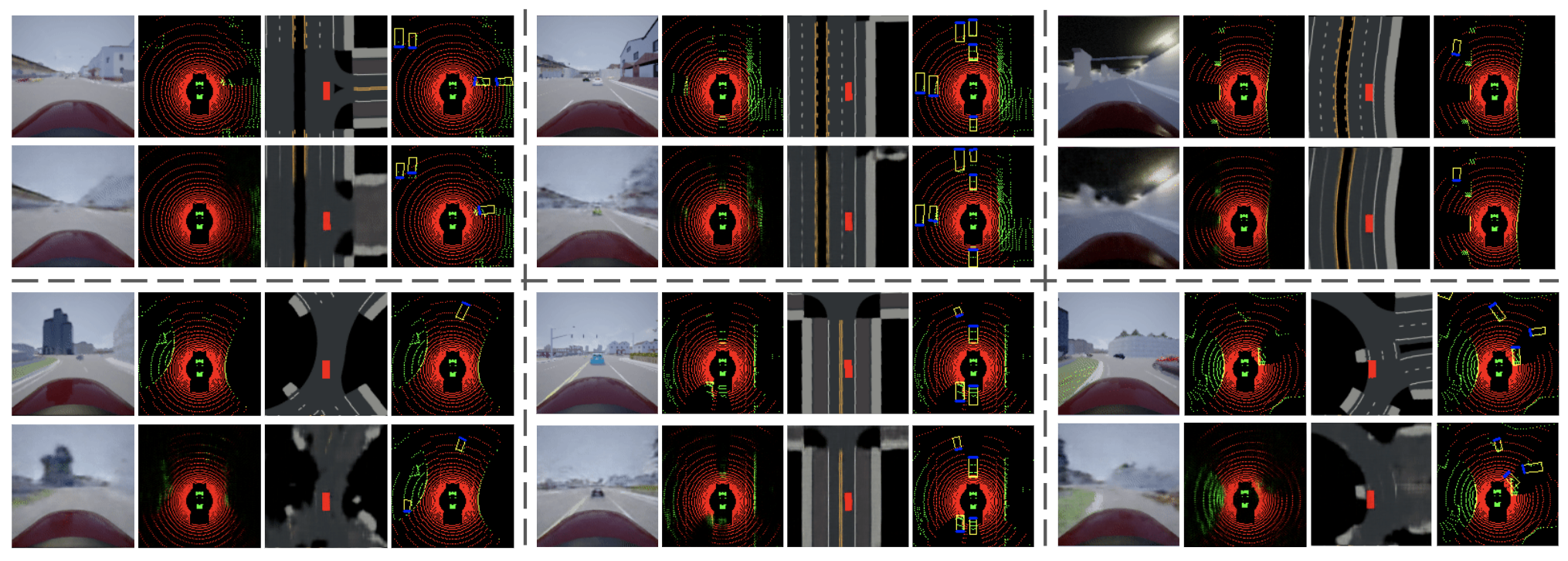}
  \caption{\label{Fig:example}Example results. The first row shows sensor inputs and ground truth, second row shows reconstructed outputs. Left to right: camera image, lidar image, local semantic roadmap, surrounding vehicle bounding boxes. Note that only camera and lidar inputs are given for the reconstruction, the ground truth semantic roadmap and bounding boxes are displayed here only for comparison.}
\end{figure*}

Fig.\ref{Fig:example} shows examples of the perception output. The first row represents the original camera and lidar inputs, as well as the ground truth local semantic roadmap and surrounding vehicle bounding boxes. The second row shows the perception output given only the historical camera and lidar inputs. We can see that the model is able to generate accurate semantic roadmap and surrounding vehicle bounding boxes, even for the occluded area.

\begin{figure}
  \centering 
  \includegraphics[width = .5\textwidth]{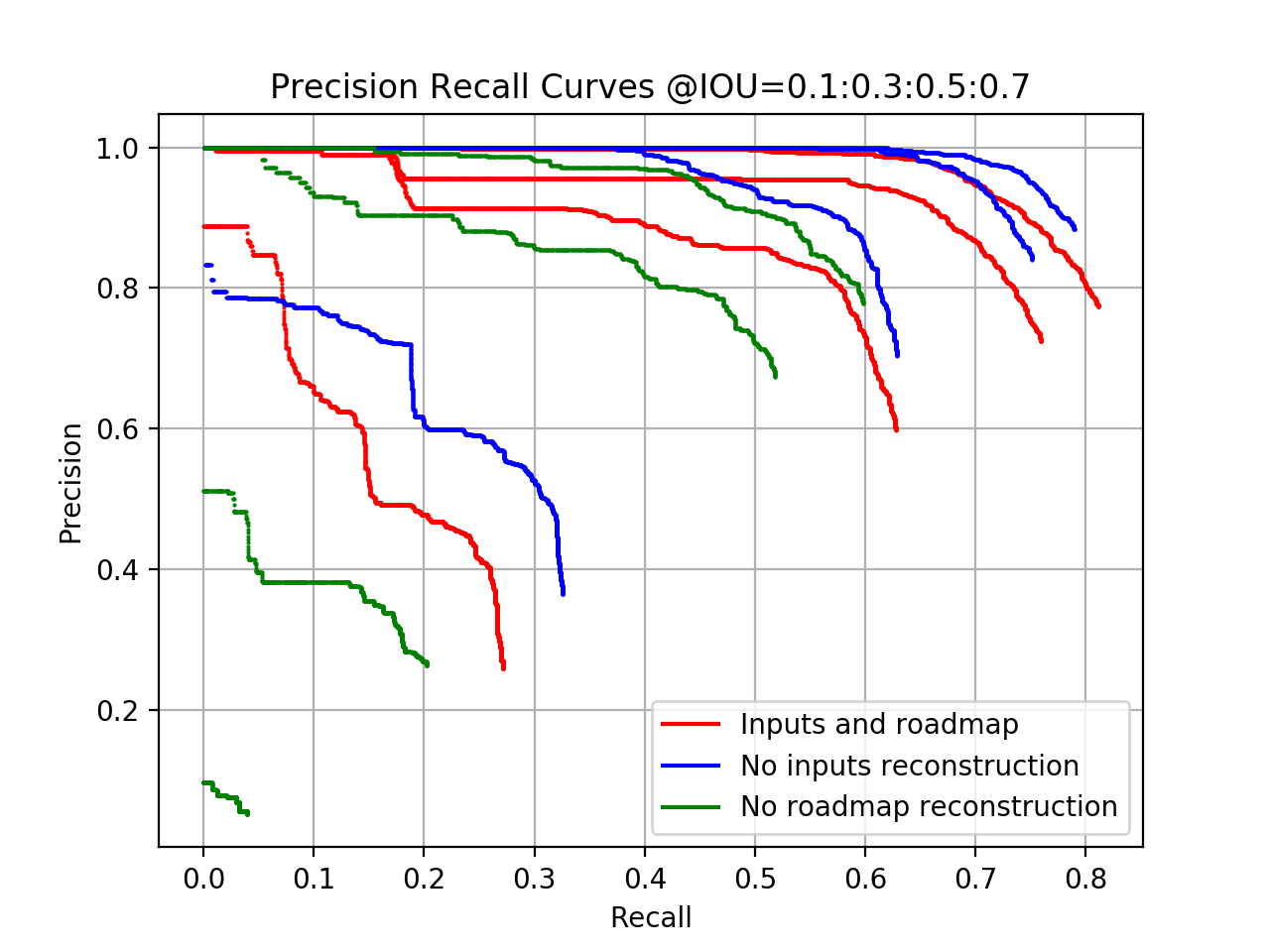}
  \caption{\label{Fig:PRC}Precision-Recall Curves for surrounding vehicles bounding boxes prediction.}
\end{figure}

We also evaluate the statistic performance of the system according to typical evaluation metrics. For surrounding vehicle bounding box prediction, we plot the Precision-Recall Curve (PRC) and then compute the Average Precision (AP) as Area Under Precision-Recall Curve (AUC)~\cite{everingham2010pascal}. The PRC and APs are computed under Intersection-Over-Union (IoU) of 0.1, 0.3, 0.5, and 0.7. Fig.\ref{Fig:PRC} shows the PRC of the methods. The APs are summarized in Table.\ref{Tab:AP}. we can see the variant that does not reconstruct the semantic roadmap has significantly worse performance than the other two variants. This shows that fusing the information of map might improve the performance of detection.

\begin{table}
\centering
\caption{Average precision for surrounding vehicles bounding boxes prediction.\label{Tab:AP}}
\resizebox{0.48\textwidth}{!}{
\begin{tabular}{c|cccc}
\hline
        & $AP_{0.1}$ & $AP_{0.3}$ & $AP_{0.5}$ & $AP_{0.7}$ \\ \hline
Inputs and roadmap & \bf{79.4}\% & 72.0\% & 56.5\% & 16.8\% \\ 
No inputs reconstruction & 78.4 & \bf{74.4\%} & \bf{61.0\%} & \bf{22.1\%} \\ 
No roadmap reconstruction & 57.4\% & 45.3\% & 7.8\% & 0.3 \% \\\hline
\end{tabular}}
\end{table}

For ego vehicle global state, we calculate its average prediction error. There are two values we care about, the location error (in meter) and heading error (in rad). Table.\ref{Tab:error} shows the evaluation results. We can see that we get an average global location error of 7.1 meters, and heading error of 0.17 rad. This is obtained purely from the raw camera and lidar sensor inputs, with no GPS or stored global map used.

\begin{table}
\centering
\caption{Average error of ego vehicle global pose estimation.\label{Tab:error}}
\resizebox{0.43\textwidth}{!}{
\begin{tabular}{c|cc}
\hline
        & Location (m) & Heading (rad) \\ \hline
Inputs and roadmap & 11.4 & 0.33  \\ 
No inputs reconstruction & 8.6 & \bf{0.17} \\ 
No roadmap reconstruction & \bf{7.1} & 0.42 \\\hline
\end{tabular}}
\end{table}

\section{Conclusion and Future Works}\label{sec:conclusion}
In this paper, we proposed and implemented an end-to-end autonomous driving perception system based on sequential latent representation learning. The system was able to replace the functionalities of the typical detection, tracking, localization and mapping subsytems with minimum human engineering efforts and without online stored maps. The method was evaluated in a realistic autonomous driving simulator, taking camera RGB image and lidar point cloud as sensor inputs. Evaluation results showed the learned model can obtain accurate surrounding vehicles' poses, local semantic roadmaps, and global ego vehicle pose based purely on historical raw sensor inputs.

Although the learned model performs reasonably well, it has a large space for improvement. The neural network architectures used in this paper are only the very basic ones, such as shallow convolutional layers. The size of input images is only $128 \times 128$, making it very hard to detect objects that are small or far away. In the future, we will deploy more advanced network architectures, and train the model on images with higher definition. We will also test this system with a downstream autonomous driving decision making system.

\bibliographystyle{ieee}
\bibliography{reference}

\end{document}